\documentclass{article}
\usepackage{spconf,amsmath,graphicx}
\usepackage{amssymb}
\usepackage{booktabs}
\usepackage{hyperref}

\usepackage{amsfonts,dsfont}
\usepackage{algorithmic}
\usepackage{graphicx,color}
\usepackage{textcomp}
\usepackage{xcolor}
\usepackage{algorithm,algorithmic}
\usepackage{arydshln}
\usepackage{float} 
\usepackage{mathtools}
\usepackage{caption}
\usepackage{subcaption}
\usepackage{cleveref}

\crefname{figure}{Fig.}{Figs.}
\crefname{table}{Tab.}{Tabs.}
\crefname{equation}{Eq.}{Eqs.}
\crefname{section}{Sec.}{Secs.}

\Crefname{figure}{Figure}{Figures}
\Crefname{table}{Table}{Tables}
\Crefname{equation}{Equation}{Equations}
\Crefname{section}{Section}{Sections}

\newcommand*{\Scale}[2][4]{\scalebox{#1}{$#2$}}%

\usepackage{xcolor}

\title{PROBABILISTIC DEEP DISCRIMINANT ANALYSIS FOR WIND BLADE SEGMENTATION}

\name{Raül~Pérez-Gonzalo$^{1,2}$, Andreas~Espersen$^{2}$ and Antonio~Agudo$^{1}$\thanks{This work has been partially supported by the project GRAVATAR PID2023-151184OB-I00 funded by MCIU/AEI/10.13039/501100011033 and ERDF, UE; and by GreenVAR project of the Fundación Ramón Areces. }}
\address{$^{1}$Institut de Robòtica i Informàtica Industrial, CSIC-UPC, Barcelona, Spain\\$^{2}$Wind Power LAB, Copenhagen, Denmark}
\begin{document}

\maketitle
\begin{abstract}
Linear discriminant analysis improves class separability but struggles with non-linearly separable data. To overcome this, we introduce Deep Discriminant Analysis (DDA), which directly optimizes the Fisher criterion utilizing deep networks. To ensure stable training and avoid computational instabilities, we incorporate signed between-class variance, bound outputs with a sigmoid function, and convert multiplicative relationships into additive ones. We present two stable DDA loss functions and augment them with a probability loss, resulting in Probabilistic DDA (PDDA). PDDA effectively minimizes class overlap in output distributions, producing highly confident predictions with reduced within-class variance. When applied to wind blade segmentation, PDDA showcases notable advances in performance and consistency, critical for wind energy maintenance. To our knowledge, this is the first application of DDA to image segmentation.
\end{abstract}


\section{Introduction}
\vspace{-0.15cm}

Linear Discriminant Analysis (LDA) is a popular linear classification technique that projects data onto a lower-dimensional space to enhance class separability. This projection is guided by the Fisher criterion, defined as the ratio of between-class to within-class variance~\cite{lda}. Apart from being computationally efficient and versatile for the multiclass setting, LDA provides robust and reliable results when handling multicollinearity and normality~\cite{bishop}.

Despite its strengths, LDA struggles with non-linearly separable data~\cite{fukunaga}. To address this, various extensions have been proposed, including uncertainty modeling for scatter matrices~\cite{scatter_lda2}, regularization techniques like Tikhonov and subspace regularization~\cite{regularized_lda3}, and advanced discriminant models with greater number of discriminants such as orthogonal~\cite{orthogonal_lda}, local~\cite{local_lda}, and heteroscedastic LDA~\cite{heteroscedastic_lda3}. LDA norm-based variants~\cite{lnorm_lda}, kernel methods~\cite{kernel_lda} and, more recently, the integration with deep learning, termed as DDA~\cite{dann,cnn_lda}, have shown promising advances in addressing LDA’s limitations.


DDA handles complex signals like images by mapping input data into a deep subspace with better class separability. It typically optimizes a Fisher-like criterion~\cite{l2_lda1,fisher_simplified,similar_lda} or an equivalent least-squares problem~\cite{l2_lda2}. However, these approaches often rely on reformulations of the Fisher criterion or require eigenvalue and scatter matrix computations, becoming computationally costly and unstable, respectively.

Conversely, modern image segmentation relies on encoder-decoder architectures~\cite{deeplabv3+,sw}, with recent developments incorporating attention mechanisms to refine feature extraction~\cite{mask2former,resnest,dsa-ilora}. Lightweight models enable efficiency in constrained environments~\cite{mobilevit,efficientformer}, while zero-shot models like SAM~\cite{sam} generalize to domain-agnostic tasks~\cite{clipseg,diffseg}.

\begin{figure}[t!]
\centering
\includegraphics[width=\linewidth]{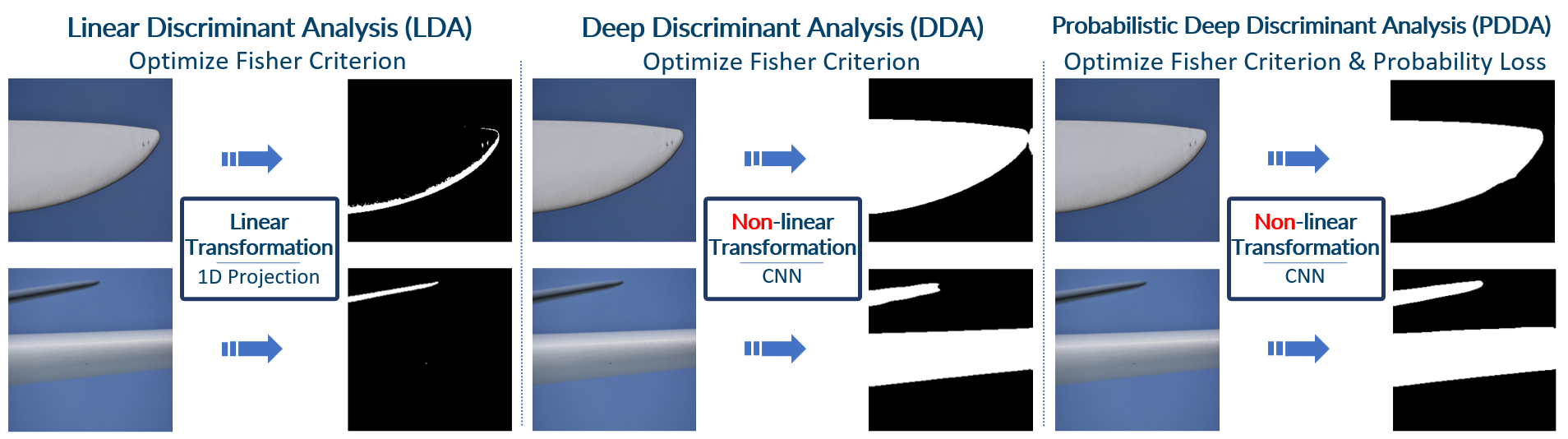} 
\vspace{-0.75cm}
\caption{\textbf{Enhancing segmentation via non-linear class separability and probability loss}. Probabilistic DDA corresponds specifically to our proposed PDDA framework.}
\label{fig:intro}
\vspace{-0.6cm}
\end{figure}

This work proposes optimizing the Fisher criterion through a Convolutional Neural Network (CNN) by adapting it to a stable loss. Novel DDA optimization is achieved by enforcing the positive class to be projected to higher values, preventing extreme gradients through the sigmoid function. This adaptation enables bounded outputs, additive relationships, and smoother gradients, facilitating robust training. We introduce two novel DDA loss functions: a log-scaled Fisher criterion and a linear decomposition. To refine the model, we combine them with a probability-based loss~\cite{focal}, forming the Probabilistic DDA (PDDA) (see \cref{fig:intro}). In this way, we provide stronger supervision for low-contrast boundaries typical of wind turbine imagery, where standard losses often struggle. 

We validate PDDA in image segmentation, becoming, to our knowledge, the first application of DDA to this task. Tailored for lightweight models with strong separation objectives, PDDA achieves accurate wind blade segmentation with limited labeled data~\cite{bunet}, capturing subtle textures and boundaries under complex conditions. This yields robust, cost-efficient solutions for the wind industry settings~\cite{PerezGonzaloIcip2024}.

\vspace{-0.1cm}
\section{Deep Discriminant Analysis}
\vspace{-0.1cm}


LDA~\cite{lda} projects data into a lower-dimensional space where class separability is maximized. By integrating class labels into its objective, LDA ensures that feature vectors $\mathbf{x}$ from different classes remain well separated after projection. For two classes, LDA computes a linear projection $y = \mathbf{w}^\top \mathbf{x}$ that best separates the background ($C_0$) and blade ($C_1$) classes, where $\mathbf{x}$ represents an RGB pixel and $\mathbf{w}$ is a learned weight vector. This projection is optimized by minimizing the inverse Fisher criterion:

\vspace{-0.5cm}
\begin{align} \label{eq:fisher}
 \mathcal L_{LDA} &=  \frac{ s^2_{C_0} + s^2_{C_1} }{(\mu_{C_0} - \mu_{C_1})^2} ~, 
 \vspace{-0.3cm}
\end{align}
where the within-class variance is measured as the sum of the projected class variances $s^2_{C_i}$, and the between-class one as the difference of the projected class means $\mu_{C_i}$, for $i\in\{0,1\}$. This leads to a generalized eigenvalue problem over $\mathbf{w}$.

Non-linear transformations enhance class separability by mapping data into a higher-dimensional space but cannot be analytically optimized as linear models. This section extends LDA to non-linear cases using a novel iterative gradient-based optimization approach that prevents gradient explosion.
\vspace{-0.25cm}
\subsection{Towards Gradient Iterative Optimization} \label{sec:nonsquared}
\vspace{-0.15cm}

We propose adopting a non-linear transformation $y = f(\mathbf{x})$ such as a neural network to separate classes. As finding the optimal non-linear solution analytically is no longer possible, we need to leverage gradient-based optimization techniques. However, directly optimizing the Fisher criterion from \cref{eq:fisher} leads to training instability, as the gradient of the between-class variance $\mathcal{G}=\nabla(\mu_{C_0} - \mu_{C_1})^2$ can point in opposite directions depending on the sampled instance:

\vspace{-0.45cm}
\begin{equation}
\Scale[0.91]{
\mathcal{G}=
  2 (\mu_{C_0} - \mu_{C_1}) (\nabla \mu_{C_0} - \nabla \mu_{C_1})
\begin{cases} 
< 0 \text{ if } \mu_{C_1} > \mu_{C_0} \\ 
> 0 \text{ if } \mu_{C_0} > \mu_{C_1}
\end{cases} \hspace{-0.3cm}.
}
\end{equation}
\vspace{-0.25cm}


To resolve this, we remove the squaring  operation and instead minimize the signed mean difference $(\mu_{C_0} - \mu_{C_1})$, while replacing the variance ratio with a product. This allows the model to learn a directional relationship where class $C_1$ is projected to higher values than $C_0$, i.e., $f(\mathbf{x}_{C_1}) > f(\mathbf{x}_{C_0})$. This reformulation enables stable gradient optimization while maintaining a discriminative objective:


\vspace{-0.5cm}
\begin{align} \label{eq:dda-div}
  \mathcal{L} =  (\mu_{C_0} - \mu_{C_1}) (s^2_{C_0} + s^2_{C_1}) ~.
\end{align}
\vspace{-0.5cm}

\vspace{-0.3cm}
\subsection{Preventing Exploding Gradients} \label{sec:sigmoid}
\vspace{-0.1cm}

In the linear setting, $\mathcal L_{LDA}$ (\cref{eq:fisher}) is scale-invariant, as rescaling the input $\mathbf{x}$ preserves the ratio of within-class to between-class variance. However, this invariance does not hold for non-linear transformations, where scaling the projected features can artificially minimize $\mathcal{L}_{LDA}$ without improving class separability. As a result, the network may increase feature magnitudes excessively, leading to exploding gradients. To counteract this, we constrain the output of the non-linear projection $f(\mathbf{x})$ using a sigmoid activation $\sigma$, bounding the output to $[0, 1]$ and stabilizing gradient updates.


Beyond numerical stability, bounding the feature space also simplifies estimation of the discriminant threshold~$\mathcal{T}$ for classifying pixels as background ($C_0$) or blade ($C_1$). A projected pixel $y = f(\mathbf{x})$ is assigned to $C_1$ if $\sigma(y) \geq \mathcal{T}$, and to $C_0$ otherwise. While in binary cross entropy the threshold is typically set near 0.5, the Fisher criterion does not originate from a probability derivation, so the optimal $\mathcal T$ must be found from the projected space, which is a priori unknown.


By restricting the output space to $[0,1]$, we enable an efficient class-threshold search to evaluate segmentation performance across candidate thresholds and select the one that maximizes class separability. 

\vspace{-0.4cm}
\subsection{Direct Deep Discriminant Analysis} \label{sec:dda_loss}
\vspace{-0.1cm}

Building on \cref{sec:nonsquared} and \cref{sec:sigmoid} over \cref{eq:fisher}, using the non-squared class mean difference for gradient stability and bounding the projected outputs via a sigmoid, we define two loss functions to enable DDA. These formulations are designed to facilitate gradient-based optimization while maintaining interpretability with respect to class separability. We introduce a parameter $\lambda_F \in \mathds{R}$ to balance the contribution of class variance terms.

\noindent
{\bf Logarithmic DDA Loss}. To balance the contribution of both class separability terms, we transform the division relationship in \cref{eq:dda-div} into an additive one by applying logarithms. Since the signed between-class variance $\mu_{C_0} - \mu_{C_1}$ lies in $[-1, 1]$, we add a constant $\epsilon = 1 + 10^{-8}$ to ensure the logarithm remains within its domain and apply it to both variance terms, ensuring that both contribute equally, writing:

\vspace{-0.55cm}
\begin{equation} \label{eq:dda_ln}
\Scale[0.95]{
\mathcal L_{DDA}^{(\ln)} = \ln \left( \epsilon + \mu_{C_0} - \mu_{C_1} \right) + \lambda_F  \ln \left( \epsilon + s^2_{C_0} + s^2_{C_1} \right) ~.
}
\end{equation}
\vspace{-0.5cm}

\noindent
{\bf Delta DDA Loss}. An alternative separability criterion is to minimize the delta difference between class variances $(s^2_{C_0} - s^2_{C_1}) - (\mu_{C_0} - \mu_{C_1})^2$~\cite{fukunaga}. Following a similar rationale, we can rearrange this criterion to adopt gradient optimization by adopting the signed between-class variance, and changing the delta difference to a sum, aiming that the blade pixels are projected to higher values than the background ones: 

\vspace{-0.55cm}
\begin{equation} \label{eq:dda_delta}
\mathcal L_{DDA}^{(\Delta)} = \left( \mu_{C_0} - \mu_{C_1} \right) + \lambda_F  \left( s^2_{C_0} + s^2_{C_1} \right) ~.
\end{equation}
\vspace{-0.55cm}

Given $m$ the binary ground-truth class, the conventional class-wise means and variances are redefined for each class $i \in \{0,1\}$ using differential formulations over the dataset~$\mathcal{D}$.

\vspace{-.65cm}
\begin{align}  
  \mu_{C_i} &= \dfrac{1}{|C_i|}\sum\limits_{y \in \mathcal{D}} m^{i}(1-m)^{(1-i)}\sigma(y) ~,   
     \\
 s^2_{C_i} &= \dfrac{1}{|C_i|-1}\sum\limits_{\mathbf{y} \in \mathcal{D}} m^i(1-m)^{(1-i)}(\sigma(y) - \mu_{C_i})^2 ~, \\
 C_i &= \sum\limits_{y \in \mathcal{D}} m^{i}(1-m)^{(1-i)} ~.
\end{align}
\vspace{-0.5cm}

\vspace{-0.45cm}
\subsection{Probabilistic Deep Discriminant Analysis}
\vspace{-0.05cm}

While class separability losses guide $f(\mathbf{x})$ to learn discriminative projections, they lack probabilistic interpretation. In contrast, binary cross-entropy reflects the negative log-likelihood of a Bernoulli distribution, enabling outputs to be treated as class probabilities. To bridge this gap, we combine class separability with a probability-based loss~\cite{focal}, which emphasizes learning from low-confidence examples.

{\bf Focal Loss}. This loss enhances the segmentation mapping by penalizing low-confidence predictions with $\gamma$ parameter. To prevent bias towards the majority class (background, in this context), class-specific weights $\alpha$ set the class balance:   

 \vspace{-0.5cm}
\begin{align} \label{eq:focal}
\mathcal L_{P} & = -\sum_{y \in \mathcal{D}} \left[ \alpha \left(1-\sigma(y)\right)^\gamma m\ln\sigma(y) \right. \nonumber \\
& \left. + (1 - \alpha) \sigma(y)^\gamma (1-m)\ln(1-\sigma(y)) \right] ~.
\end{align}
\vspace{-0.45cm}

\noindent
{\bf Probabilistic DDA Loss}. We define the final loss as a linear combination of the DDA loss (with the superscript indicating the chosen DDA) and Focal Loss, weighted by $\lambda_P$:

\vspace{-0.35cm}
\begin{equation}
\mathcal L_{PDDA} = \mathcal L_{P} + \lambda_P \mathcal L_{DDA} ~.
\end{equation}
 \vspace{-0.45cm}



\vspace{-0.4cm}
\subsection{Implementation Details} \label{sec:implementation}
\vspace{-0.05cm}

Input images are resized to $256 \times 256$, min-max normalized, and augmented with flipping and cropping. We adopt U-Net~\cite{unet} as the backbone architecture. We use Adam optimizer with an initial learning rate of $10^{-4}$ and batch size 8. A custom scheduler reduces the learning rate after three validation plateaus. The variance balance parameters are $\lambda_F = 0.4$ for $\mathcal L_{DDA}^{(\Delta)}$ and $\lambda_F = 0.9$ for $\mathcal L_{DDA}^{(\ln)}$. Focal loss parameter are fine-tuned over the validation set to $\gamma = 2$ and $\alpha = 0.25$.




    



\vspace{-0.25cm}
\section{Experimental Results} \label{sec:results}
\vspace{-0.15cm}


This section compares LDA and DDA performance and, then, presents both quantitative and qualitative evaluations to showcase the effectiveness of PDDA, highlighting the methods' robustness by comparing results across different windfarms. The wind blade dataset employed is taken from~\cite{bunet}.




\begin{table}[t!]
\centering
\resizebox{0.93\linewidth}{!} {
\begin{tabular}{lcccccccccc}
\toprule
     \multicolumn{1}{c}{Method} & $\lambda_P$  & \multicolumn{1}{c}{Accuracy}  & \multicolumn{1}{c}{F1} & \multicolumn{1}{c}{mIoU} & $\mu_{C_1}-\mu_{C_0}$ & $s^2_{C_0}$ & $s^2_{C_1}$\\
    \multicolumn{1}{c}{} & & {[\%]}  & {[\%]}  & {[\%]} & $\uparrow$ & $\downarrow$ & $\downarrow$ \\
    \midrule

   LDA  & - & 72.59 & 58.46 & 53.67 & - & - & - \\

   DDA$^{(\Delta)}$ & -  & 94.06 & 91.52 & 87.11 & .81 & .033 & .14 \\ 
   DDA$^{(\ln)}$ & - & 95.50 & 91.59 & 88.70 & .83 & .023 & .13 \\ \hdashline[0.5pt/2pt] 

   $\mathcal{L}_P$ & -  & 96.28 & 94.05 & 90.91 & .69 & .007 & .05  \\  
   
   PDDA$^{(\Delta)}$ & .0001 & 94.27 & 85.12 & 86.34 & .47  &  .007  & .02  \\
   PDDA$^{(\Delta)}$ & .001 & 95.15 & 93.23 & 90.12 & .57 & .008  & .04 \\
   PDDA$^{(\Delta)}$ & .01 & 96.82 & 94.57 & 91.87 & .66 & .010 & .07 \\
   PDDA$^{(\Delta)}$ & .1 & 97.10 &  95.42 & 93.05 & .77 & .010 & .10 \\
   PDDA$^{(\Delta)}$ & 1 & \textbf{97.62} & \textbf{95.53} & \textbf{93.29} & .77 & .007 & .10 \\
   PDDA$^{(\Delta)}$ & 2 & 95.82 & 91.47 & 90.07 & .82 & .016 & .09 \\ \hdashline[0.5pt/2pt] 
   
   PDDA$^{(\ln)}$ & .0001 & 96.51 & 93.04 & 90.85 & .49 & .010 & .05 \\ 
   PDDA$^{(\ln)}$ & .001 & 96.56 & 93.48 & 91.57 & .54 & .009 & .05 \\ 
   PDDA$^{(\ln)}$ & .01 & 97.19 & 94.80 & 92.32 & .64 & .014 & .06 \\ 
   PDDA$^{(\ln)}$ & .1 & \textbf{97.97} & \textbf{96.28} & \textbf{94.34} & .76 & .012 & .11 \\ 
   PDDA$^{(\ln)}$ & 1 & 96.77 & 94.82 & 92.23 & .74 & .009 & .17 \\ 
   PDDA$^{(\ln)}$ & 2 & 96.00 & 90.70 & 89.73 & .80 & .016 & .13 \\ 
\bottomrule
\end{tabular}
}
\vspace{-0.3cm}
\caption{\textbf{Quantitative comparison} of $\mathcal{L}_P$, DDA and PDDA.}  
\label{tab:dist}
\vspace{-0.4cm}
\end{table}

\begin{figure}[t!]
\hspace{0.2cm}\begin{tabular}{c}
\includegraphics[width=0.88\linewidth]{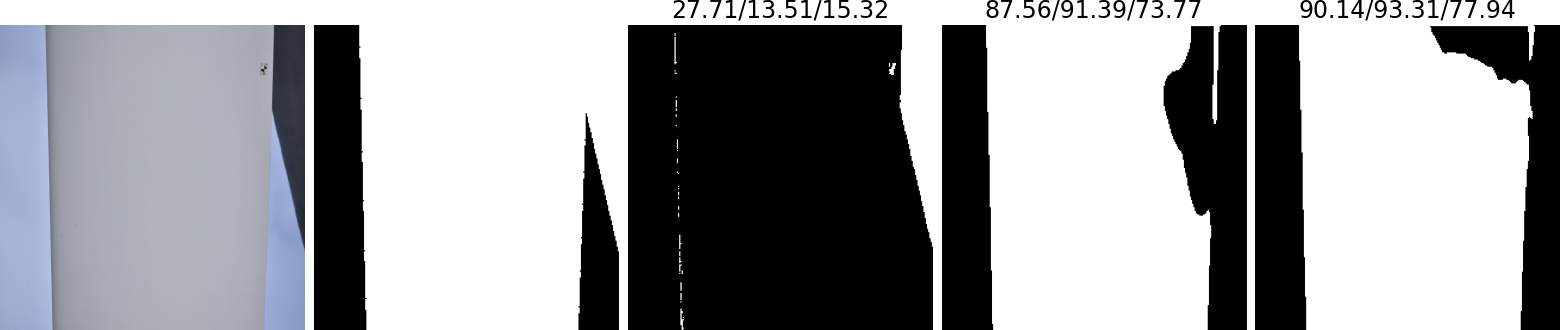}\\ \vspace{-0.05cm}
\includegraphics[width=0.88\linewidth]{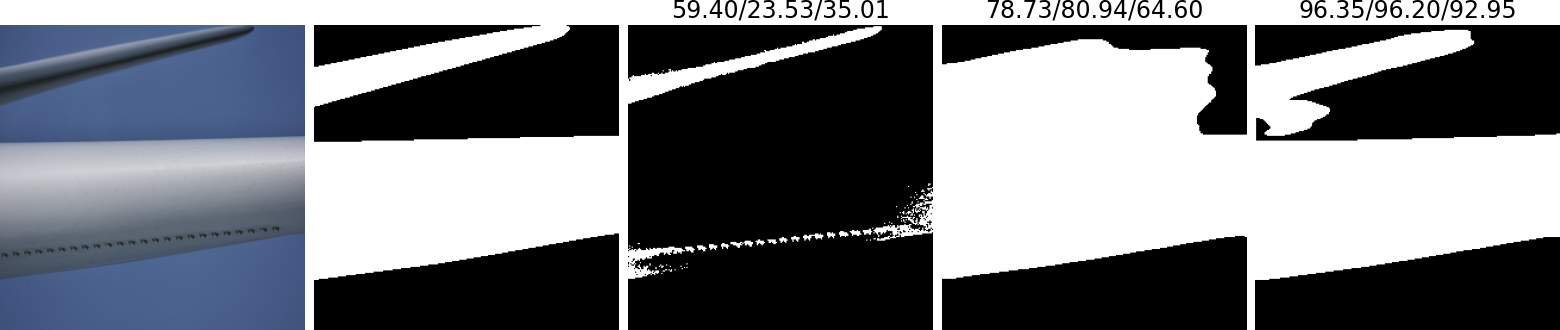}\\ \vspace{-0.05cm}
\includegraphics[width=0.88\linewidth]{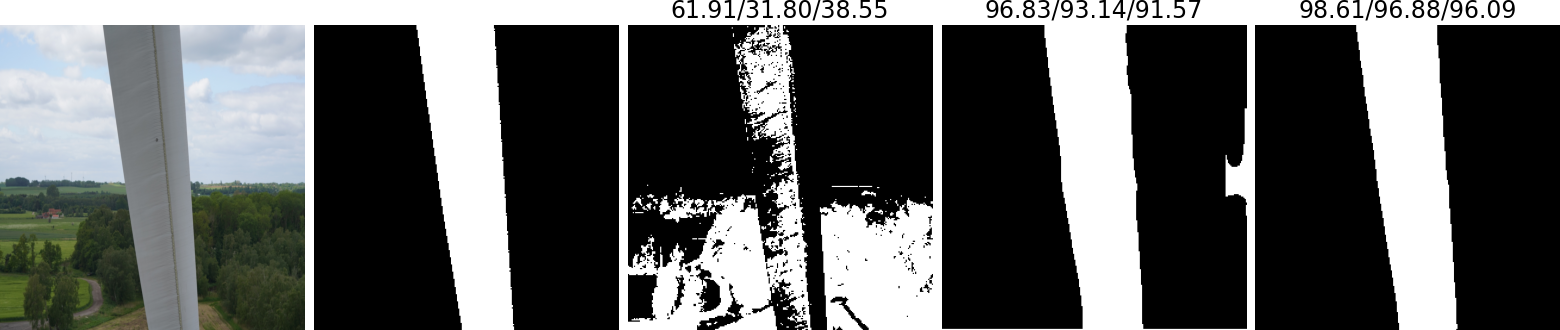}\\ \vspace{-0.05cm}
\includegraphics[width=0.88\linewidth]{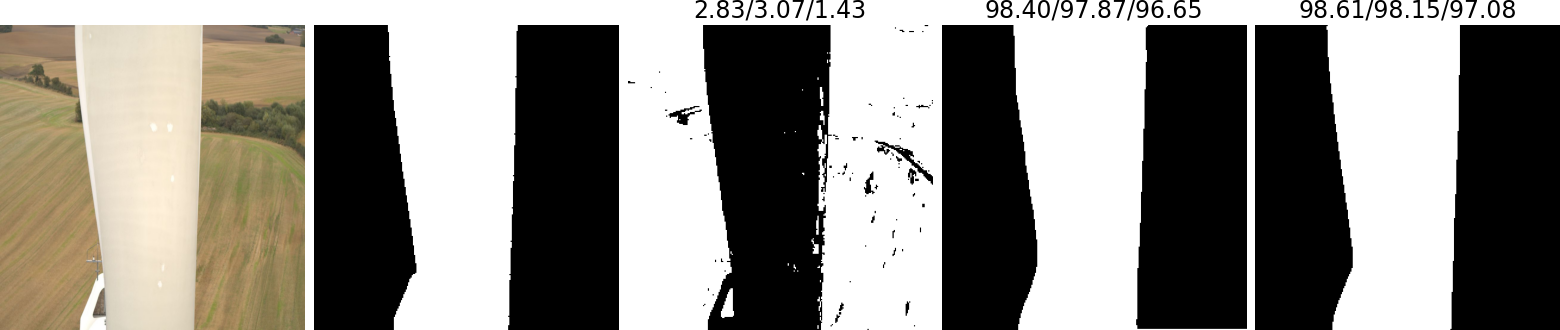}

\end{tabular}
\vspace{-0.45cm}
\caption{\textbf{Qualitative comparison of LDA and DDA.} From left-to-right: input image, ground-truth mask, LDA, DDA$^{(\Delta)}$ and DDA$^{(\ln)}$ estimations with accuracy, F1, and mIoU above.}
\label{fig:seg-lda-visual}
\vspace{-0.65cm}
\end{figure}

\vspace{-0.25cm}
\subsection{Direct Deep Discriminant Analysis}
\vspace{-0.1cm}

We first explore the benefits of using a non-linear transformation as a deep network for optimizing discriminant analysis. As shown in \cref{tab:dist}, DDA significantly outperforms LDA. Both optimization objectives, $L_{DDA}^{(\Delta)}$ in \cref{eq:dda_delta} and $L_{DDA}^{(\ln)}$ in \cref{eq:dda_ln}, yield comparable gains over LDA, with accuracy improving by about 20\% and recall doubling. A dummy classifier predicting only the majority class (background) already achieves 69\% accuracy, close to LDA’s 72.59\%, confirming that the blade segmentation task is non-linearly separable. From a qualitative perspective, \cref{fig:seg-lda-visual} illustrates that LDA often fails in complex backgrounds or only partially segments the blade, whereas DDA correctly identifies the blade region in most cases (last two rows). Remaining challenges (top two rows) are further addressed by PDDA in \cref{sec:qualitative}.


\vspace{-0.2cm}
\subsection{Probabilistic Deep Discriminant Analysis}
\vspace{-0.15cm}

Detailed insights into the adoption of PDDA are provided in \cref{tab:dist}, including performance metrics and class variance breakdowns. Both $\mathcal{L}_P$ and DDA achieve remarkable segmentation performance, but $\mathcal{L}_P$ overperforms DDA, because it follows the state-of-the-art BU-Net~\cite{bunet} designed for blade wind segmentation. Although the between-class variance ($\mu_{C_1} - \mu_{C_0}$) is not as pronounced as with DDA, optimizing $\mathcal{L}_P$ results in significantly lower within-class variances ($s^{2}_{C_i}$ for $i \in \{0, 1\}$), even though it does not directly minimize them. By balancing the probability and DDA terms with $\mathcal{L}_P$, PDDA retains the high between-class variance from DDA and the low within-class variances from $\mathcal{L}_P$, leading to top-performing segmentation metrics. Note that PDDA$^{(\Delta)}$ is balanced with $\lambda_P = 1$, whereas PDDA$^{(\ln)}$ requires $\lambda_P = 0.1$ due to $10 \cdot \mathcal L_{DDA}^{(\Delta)} \approx \mathcal L_{DDA}^{(\ln)}$ over the range $[0, 1]$.

\vspace{-0.4cm}
\subsection{Quantitative Evaluation}
\vspace{-0.15cm}

\Cref{tab:seg-unet-compare} provides a quantitative comparison of several state-of-the-art segmentation methods, trained from scratch following the data pipeline and augmentation from \cref{sec:implementation}. The U-Net~\cite{unet}, trained with binary cross-entropy, is used as the baseline, given its role as our backbone architecture. While it achieves respectable accuracy, its recall rate of 68.93\% reveals a notable shortfall in capturing the blade region.

More advanced architectures, such as Mask2Former~\cite{mask2former}, markedly improve on U-Net, with stronger results across multiple metrics, particularly in recall. Zero-shot models like SAM~\cite{sam} also achieve impressive quantitative scores. BU-Net~\cite{bunet} surpasses prior state-of-the-art methods as it is tailored to our specific dataset. Still, PDDA$^{(\Delta)}$ and PDDA$^{(\ln)}$ stand out by delivering top-tier performance compared to these segmentation algorithms. Notably, PDDA$^{(ln)}$ achieves the highest scores across all metrics except precision, where BU-Net~\cite{bunet} leads. This underscores PDDA$^{(ln)}$’s superiority, with 97.97\% accuracy, 96.28\% F1-score and 94.34\% mIoU.

Moreover, PDDA’s inference computational cost is comparable to the well-studied U-Net~\cite{unet}, offering both an efficient and highly effective segmentation solution.

\begin{table}[t!]
\centering
\resizebox{\linewidth}{!} {
\begin{tabular}{lccccccccc}
\toprule
     \multicolumn{1}{c}{Method} & \multicolumn{1}{c}{Accuracy}  & \multicolumn{1}{c}{Precision} & \multicolumn{1}{c}{Recall} & \multicolumn{1}{c}{F1} & \multicolumn{1}{c}{mIoU} & \multicolumn{1}{c}{IoU$_{C_0}$} & \multicolumn{1}{c}{IoU$_{C_1}$} \\
    \multicolumn{1}{c}{} & {[\%]} & {[\%]} & {[\%]}  & {[\%]}  & {[\%]} & {[\%]}  & {[\%]}  \\
    \midrule
    U-Net~\cite{unet} & 86.24 & 95.51 & 68.93 & 77.95 & 75.94 & 80.31 & 71.57 \\
    DeepLabv3+~\cite{deeplabv3+}  & 94.14 & 96.36 & 87.38 & 89.03 & 87.47 & 90.31 & 84.62 \\  
    SW~\cite{sw} & 93.48 & 93.57 & 91.71 & 91.37 & 87.44 & 88.64 & 86.23 \\ 
    ResNeSt~\cite{resnest} & 94.23 & 96.84 &91.47 & 92.77 & 89.63 & 90.40 & 88.86 \\ 
    SAM~\cite{sam} & 94.36 & 97.29 & 91.22 & 92.60 & 91.66 & 92.31 & 91.01  \\ 
    CLIPSeg~\cite{clipseg} &  82.70 & 77.02 & 75.52 & 74.29 & 75.09 & 80.16 & 70.02 \\
    DiffSeg~\cite{diffseg} & 96.37 & 82.08 & 89.74 & 85.73 & 86.40 & 91.66 & 81.13 \\
    EfficientFormer~\cite{efficientformer} & 96.42 & 95.47 &  93.63 & 94.55 & 93.51 & 94.02 & 92.99 \\ 
    MobileViT~\cite{mobilevit} & 96.14 & 95.44 & 93.33 & 94.38 & 93.47 & 94.06 & 92.88\\
    Mask2Former~\cite{mask2former} & 96.68 & 95.63 & 93.89 & 94.76 & 93.72 & 94.51 & 92.93\\
    BU-Net~\cite{bunet} & 97.39 & \textbf{99.42} & 93.35 & 95.73 & 93.80 & 94.70 & 92.90\\ 
   PDDA$^{(\Delta)}$ & 97.62 & 94.29 & 97.79 & 95.53 & 93.29 & 94.36 & 92.21 \\ 
   PDDA$^{(\ln)}$ & \textbf{97.97} & 94.66 & \textbf{98.71} & \textbf{96.28} & \textbf{94.34}  & \textbf{95.28} & \textbf{93.40} \\ 

\bottomrule
\end{tabular}
}
\vspace{-0.3cm}
\caption{\textbf{Quantitative comparison on blade segmentation} with respect to state-of-the-art methods. } 
\vspace{-0.45cm}
\label{tab:seg-unet-compare}
\end{table}

\begin{figure}[t!]
\hspace{-0.1cm}\begin{tabular}{c}
\includegraphics[width=0.99\linewidth]{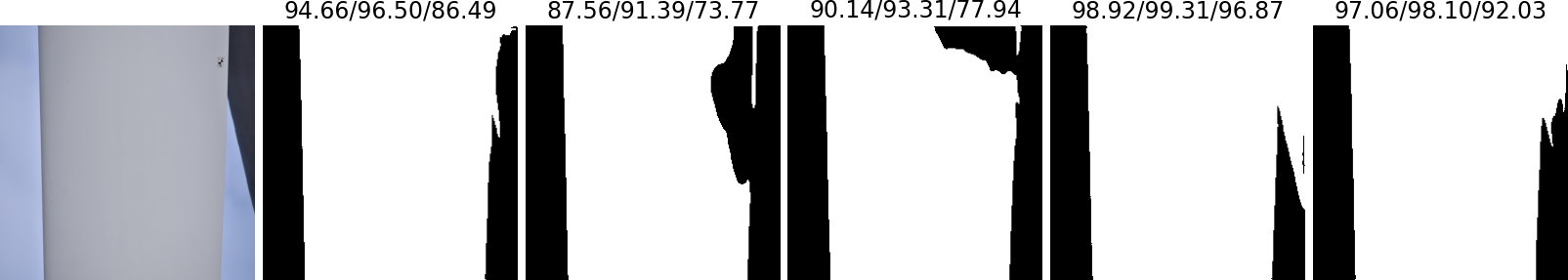}\\
\includegraphics[width=0.99\linewidth]{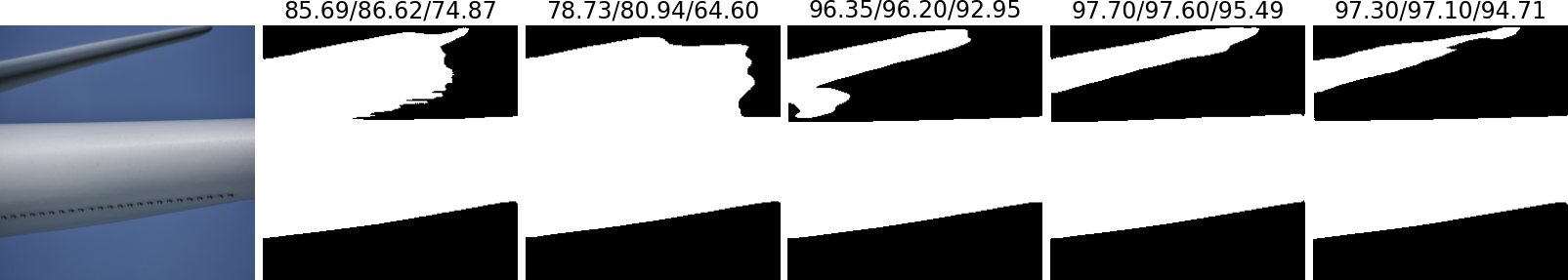}\\
\includegraphics[width=0.99\linewidth]{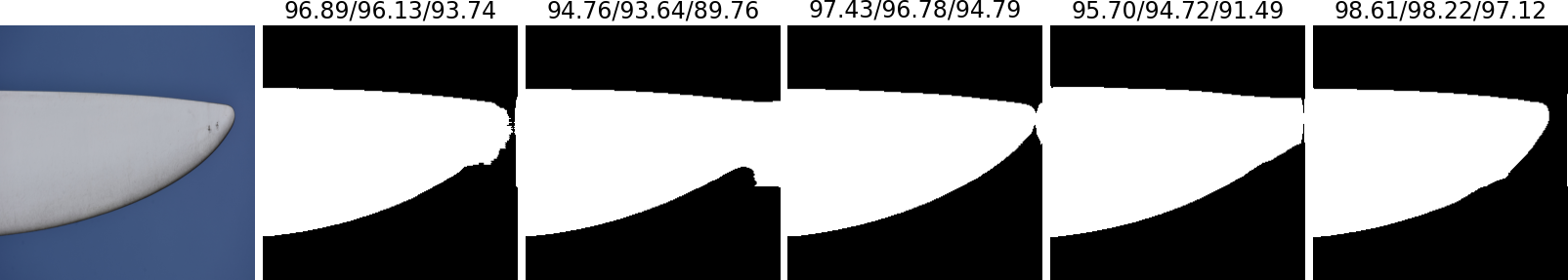}\\
\includegraphics[width=0.99\linewidth]{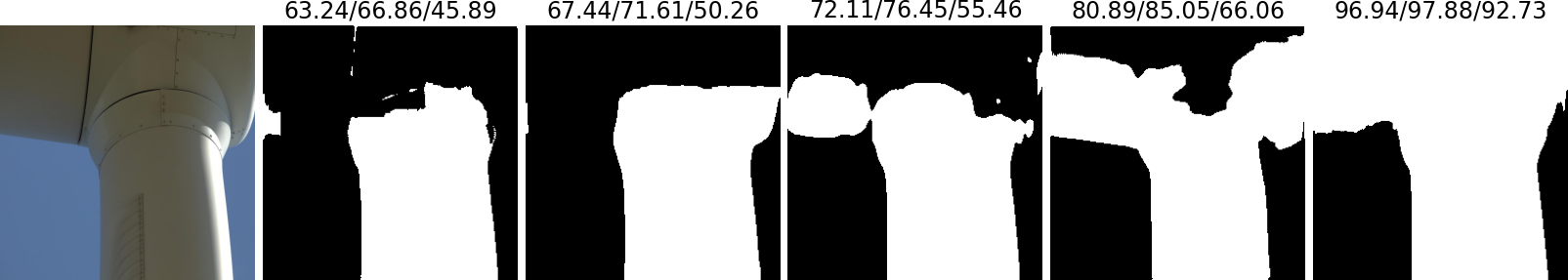}\\
\includegraphics[width=0.99\linewidth]{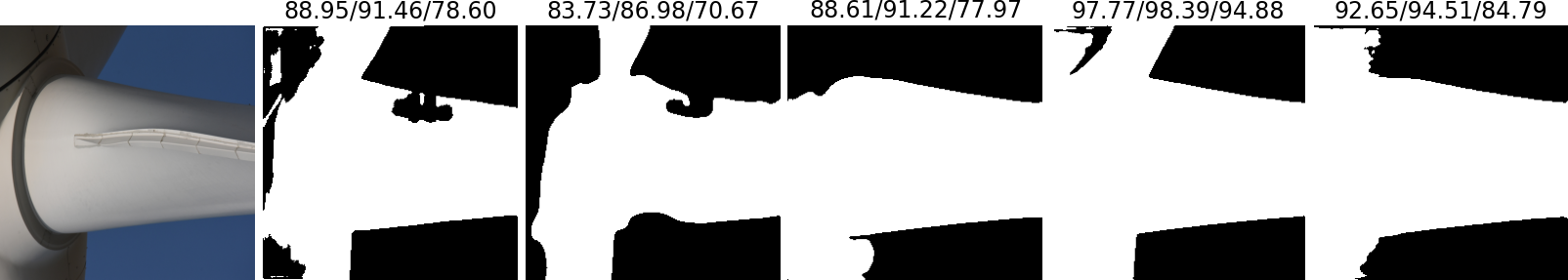}
\end{tabular}
\vspace{-0.45cm}
\caption{\textbf{Qualitative comparison of $\mathcal{L}_P$, DDA and PDDA.} From left to right columns: input image, $\mathcal{L}_P$, DDA$^{(\Delta)}$, DDA$^{(\ln)}$, PDDA$^{(\Delta)}$ and PDDA$^{(\ln)}$ estimations with accuracy, F1-score and mIoU above.}
\label{fig:seg-visual}
\vspace{-0.4cm}
\end{figure}

\vspace{-0.25cm}
\subsection{Qualitative Evaluation} \label{sec:qualitative}
\vspace{-0.1cm}

Despite the high performance metrics, \cref{fig:seg-visual} exemplifies some images where probabilistic and DDA are necessary in conjunction to accomplish high-quality segmentation. This complexity arises from limited data and complex blade shapes, high color transitions within regions, and shadows partially obscuring blade areas. Combining DDA with probability-based loss improves accuracy, overperforming models trained solely with DDA or probability losses. In particular, we can observe that PDDA excels other models in capturing complex boundaries in the hub or additional wind turbine blades behind the primary salient blade.

\vspace{-0.25cm}
\subsection{Generalization Across Windfarms}
\vspace{-0.1cm}

\begin{figure}[t!]
\centering
\hspace{-0.2cm}\includegraphics[width=1.0\linewidth]{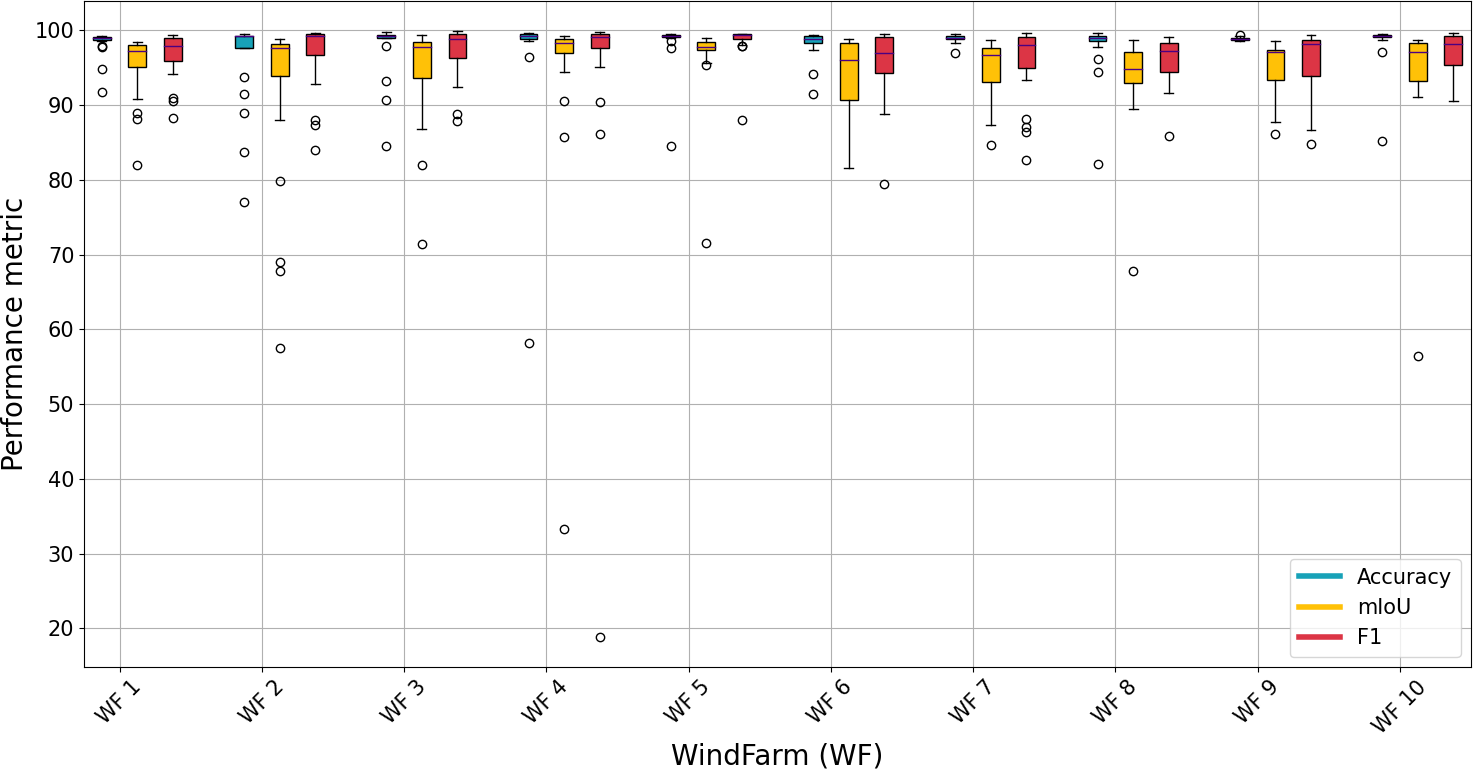} 
\vspace{-0.3cm}
\caption{\textbf{Boxplot results of PDDA$^{(\ln)}$ over the test set for each windfarm} to study the robustness of our algorithm.} 
\label{fig:windfarm} 
\vspace{-0.6cm}
\end{figure}

To assess PDDA's generalizability and robustness, the test set comprises 20 randomly selected images from 10 distinct windfarms and inspection campaigns~\cite{bunet}. \Cref{fig:windfarm} presents the performance distribution per windfarm. The results indicate consistently high performance across all windfarms, confirming the robustness and generalization capabilities of our approach in effectively segmenting unseen test images. 

\vspace{-0.15cm}
\section{Conclusion}
\vspace{-0.2cm}


LDA is a foundational linear classifier, but its reliance on linear separability limits broader applicability. We presented DDA, which leverages neural networks to directly optimize the Fisher criterion in a stable manner without costly eigen-decomposition. Through novel loss functions, DDA improves class separability and segmentation quality and provides stronger supervision for low-contrast boundaries common in wind blade imagery. We also introduced PDDA, which combines discriminant learning with probability-based loss to reduce class uncertainty and within-class variance. Experiments on wind blade segmentation, a domain with scarce labels and fine-grained structures, demonstrate PDDA’s effectiveness and practical impact. By unifying discriminant analysis with probabilistic learning in lightweight models, PDDA provides robust, cost-efficient segmentation and opens new directions for broader industrial applications.

\bibliographystyle{IEEEbib}
\bibliography{strings}

\end{document}